\documentclass[10pt,a4paper]{article}

\usepackage[utf8]{inputenc}
\usepackage[T1]{fontenc}
\usepackage{lmodern}
\usepackage[margin=2.5cm]{geometry}
\usepackage{booktabs}
\usepackage{tabularx}
\usepackage{hyperref}
\usepackage{enumitem}
\usepackage{amsmath}
\usepackage{authblk}
\usepackage{tikz}
\usetikzlibrary{shapes,arrows,positioning,shadows}

\hypersetup{
    colorlinks=true,
    linkcolor=blue,
    citecolor=blue,
    urlcolor=blue
}

\title{The Convergence of Schema-Guided Dialogue Systems and the Model Context Protocol: A New Paradigm for Agentic Interoperability}
\author{Andreas Schlapbach\footnote{\copyright\ 2026 Andreas Schlapbach. Licensed under \href{https://creativecommons.org/licenses/by-sa/4.0/}{CC BY-SA 4.0}. Research supported by SBB-IT. The views expressed in this paper are the author's own and do not necessarily reflect the views or policies of SBB. ORCID: 0009-0006-2329-2626.}}
\affil{SBB (Swiss Federal Railways), SBB IT, Bern, Switzerland \\
\texttt{schlpbch@gmail.com} \\ }
\date{March 4th, 2026}

\begin{document}

\maketitle

\begin{abstract}
This paper establishes a fundamental convergence: Schema-Guided Dialogue (SGD) and the Model Context Protocol (MCP) represent two manifestations of a unified paradigm for deterministic, auditable LLM-agent interaction. SGD, designed for dialogue-based API discovery (2019), and MCP, now the de facto standard for LLM-tool integration, share the same core insight---that schemas can encode not just tool signatures but operational constraints and reasoning guidance. By analyzing this convergence, we extract five foundational principles for schema design: (1) Semantic Completeness over Syntactic Precision, (2) Explicit Action Boundaries, (3) Failure Mode Documentation, (4) Progressive Disclosure Compatibility, and (5) Inter-Tool Relationship Declaration. These principles reveal three novel insights: first, SGD's original design was fundamentally sound and should be inherited by MCP; second, both frameworks leave failure modes and inter-tool relationships unexploited---gaps we identify and resolve; third, progressive disclosure emerges as a critical production-scaling insight under real-world token constraints. We provide concrete design patterns for each principle. These principles position schema-driven governance as a scalable mechanism for AI system oversight without requiring proprietary system inspection---central to Software 3.0.
\end{abstract}


\section{Introduction}
\label{sec:introduction}

The emergence of large language models (LLMs) has fundamentally changed how we approach artificial intelligence. Unlike traditional expert systems or narrow task-specific models, modern LLMs possess general reasoning capabilities that enable them to understand and respond to a vast range of inputs. These capabilities are grounded in few-shot and in-context learning~\cite{gpt3, icl-foundations, incontext-learning-review}, where models adapt to new tasks through examples provided in the input context rather than requiring retraining. However, this generality comes with a critical limitation: LLMs are trained on static datasets and have no native ability to perceive the real world, access external information, or take actions in response to user requests. To bridge this gap, the research and practitioner communities have developed methods and standards for connecting language models to external tools, APIs, and data sources.

The history of task-oriented dialogue systems---conversational AI systems designed to help users accomplish specific goals---illustrates this evolution. Early systems like MultiWOZ\cite{multiwoz} relied on predefined ontologies: a fixed set of domains, intents, and slots that the dialogue system could handle. When a new service or API needed to be integrated, the system required extensive retraining. This ``ontology bottleneck'' became a fundamental limitation on scalability. The Schema-Guided Dialogue (SGD) framework, introduced by Google in 2019~\cite{sgd-arxiv}, demonstrated that a single dialogue model could generalize zero-shot to new APIs by understanding natural language schema descriptions at runtime. This breakthrough in data-driven, service-agnostic dialogue was a watershed moment in conversational AI.

Yet SGD remained largely a research contribution. The practical, real-world adoption of schema-guided reasoning required a standardized protocol---not just a dataset and algorithms, but a specification that any LLM application could use to connect with any external service. In late 2024, Anthropic introduced the Model Context Protocol (MCP)~\cite{mcp-spec}, an open standard designed to solve the ``N-to-M integration problem'': instead of each AI application building custom integrations with each tool or service, MCP provides a universal protocol that hosts and servers can use to communicate in a standardized manner. MCP operationalizes the principles of schema-guided reasoning, bringing them from research into practical deployment.

This paper explores the convergence of these two paradigms. We argue that SGD and MCP are not separate developments but rather different manifestations of a unified principle: that autonomous agents can dynamically discover and reason about services through machine-readable schema descriptions, without requiring retraining or hardcoded knowledge. We trace how this principle evolved from theoretical foundations (transformers, few-shot learning, chain-of-thought reasoning) through research breakthroughs (SGD datasets and models) to current industry practice (MCP deployments). We examine the architectural patterns that enable this convergence, the benchmarks that measure progress, the optimization strategies that make it practical, and the security considerations that make it safe.

\subsection*{Practical Validation: Federated Agent Ecosystem}

This analysis is grounded in experience with an ecosystem of over 10 agents. The agents are each a domain expert, following domain-driven design \cite{vernon2016ddd}. These agents are intelligently federated, coordinating through dynamically discovered schema relationships and MCP protocols rather than rigid orchestration logic. This federated architecture has revealed critical gaps between the theoretical elegance of SGD principles and the practical requirements of production agent systems: over 1,000 tool-to-tool dependencies must be managed, action boundaries must be enforced deterministically, and failure modes must be recoverable without human intervention. The analysis presented in this paper draws directly from these operational constraints, validating both the soundness of SGD's original principles and the critical standardization gaps that MCP must address to reliably scale beyond ad-hoc deployments. The ecosystem's design principles and its intelligent federation will be described in detail in a future paper.

The structure of this paper is as follows. Section~\ref{sec:foundations-sgd} establishes the foundations of the SGD paradigm, examining the ontology bottleneck and how SGD solved it. Section~\ref{sec:structural-components} describes the structural components of service schemas in SGD. Section~\ref{sec:data-collection} discusses data collection methods for SGD systems. Section~\ref{sec:mcp-standard} introduces the Model Context Protocol, its architecture, and its primitives. Section~\ref{sec:sgd-mcp-mapping} maps SGD principles to MCP concepts, showing their structural correspondence. Section~\ref{sec:dynamic-discovery} examines dynamic discovery and the ``USB-C for AI'' metaphor. Section~\ref{sec:state-tracking} discusses state tracking and multi-turn reasoning in agentic systems. Section~\ref{sec:compass} introduces the COMPASS architecture for managing long-horizon tasks. Section~\ref{sec:benchmarking} surveys benchmarking approaches for evaluating agentic performance. Section~\ref{sec:optimization} addresses the token bloat problem and optimization strategies. Section~\ref{sec:security} analyzes security and trust in MCP ecosystems. Section~\ref{sec:network-management} presents a case study in network management. Section~\ref{sec:mathematical-foundations} covers mathematical foundations for ensemble agent decisions. Section~\ref{sec:schema-design} proposes five foundational principles for designing schemas that serve LLM agents effectively. Finally, Section~\ref{sec:conclusion} concludes with a vision of Software~3.0 and identifies open research challenges.

\section{Foundations of the Schema-Guided Dialogue Paradigm}
\label{sec:foundations-sgd}

The historical limitations of task-oriented dialogue systems were largely
defined by the ``ontology bottleneck.'' Early systems were trained on datasets
like MultiWOZ, which relied on a predefined set of domains, intents, and slots.
When a new service---such as a specific restaurant booking API or a new flight
carrier---needed to be integrated, the system typically required new annotated
training data to recognize the specific parameters of that service. The
Schema-Guided Dialogue (SGD) dataset was developed specifically to overcome
these scaling challenges by introducing a benchmark for virtual assistants that
must support a large and constantly changing number of services~\cite{sgd-arxiv, sgd-aaai, lee2022sgd}.

The SGD dataset contains over 20,000 multi-domain conversations spanning 20
distinct domains, including banking, events, media, and travel~\cite{sgd-github}. Its core
innovation is the requirement for models to make predictions over a dynamic set
of intents and slots provided as input via their natural language descriptions.
This setup allows a single unified model to handle services it has never
encountered during training, facilitating zero-shot generalization~\cite{gpt3}. The challenge of joint intent detection and slot filling---recognizing both what the user wants (intent) and the relevant parameters (slots)---has been extensively studied~\cite{intent-slot-joint, intent-detection-survey}. Modern neural architectures like BERT~\cite{bert} have proven effective at this task, learning to
interpret schema descriptions and ground reasoning in contextual information. The
importance of this shift cannot be overstated, as it mirrors the way human users
discover new application features through documentation rather than retraining
their linguistic faculties.

\section{Structural Components of Service Schemas}
\label{sec:structural-components}

In the SGD framework, the interface of an API is defined through a structured
JSON object known as a schema~\cite{sgd-arxiv, sgd-gem}. This schema acts as the ``contract'' between the
dialogue system and the underlying service. It encapsulates not only the
technical requirements of the API but also the semantic context necessary for a
language model to understand the user's intent. Table~\ref{tab:schema-fields} presents the key fields that define an SGD schema.

\begin{table}[ht]
\centering
\small
\caption{Schema field definitions in the SGD framework.}
\label{tab:schema-fields}
\vspace{0.2cm}
\begin{tabularx}{\textwidth}{@{}l X X@{}}
\toprule
\textbf{Field} & \textbf{Definition and Purpose} & \textbf{Metadata Requirements} \\
\midrule
\texttt{name} & The unique identifier for the intent or function (e.g., GetWeather). & Alphanumeric string \\
\texttt{description} & A natural language explanation of what the intent accomplishes. & Descriptive prose \\
\texttt{is\_transactional} & A boolean indicating if the call involves a permanent state change (e.g., a purchase). & Boolean \\
\texttt{required\_slots} & Parameters that must be collected before the API call can be executed. & List of slot names \\
\texttt{optional\_slots} & Parameters that may be provided, often with default values. & Map of name to default \\
\texttt{slot\_values} & Possible values for categorical slots, often with linguistic variations. & List of strings \\
\bottomrule
\end{tabularx}
\end{table}

The dataset utilizes a ``frame-based'' representation for dialogue state tracking,
where each frame corresponds to a single service present in the dialogue. This
modularity allows the system to track active intents, requested slots, and the
values assigned to those slots across multiple turns, even as the user switches
between different domains like booking a hotel and then checking the weather in
the destination city.

\section{Data Collection and Simulation Mechanisms}
\label{sec:data-collection}

The creation of the SGD dataset relied on a novel dialogue simulator rather than
the traditional, labor-intensive Wizard-of-Oz (WoZ) approach \cite{multiwoz}. While WoZ setups are often prone to human annotation errors and are difficult to scale, the SGD
simulator generates dialogue outlines over arbitrary combinations of APIs using
a probabilistic automaton~\cite{sgd-arxiv}. This automaton is designed to capture a wide variety of dialogue trajectories, ensuring that the resulting dataset is representative
of real-world complexity. The simulator consists of two agents playing the roles
of the user and the assistant, interacting through a finite set of dialogue
acts. These structured outlines are then converted into natural language
utterances using action templates. Finally, crowd-workers paraphrase these
templatized utterances to make the dialogue flow more naturally and coherently,
without needing to provide extra annotations. This process ensures high-quality,
consistent annotations while significantly reducing the cost and time of data
collection.

\section{The Model Context Protocol: An Operational Standard for Interoperability}
\label{sec:mcp-standard}

While the SGD paradigm provided the theoretical framework for schema-guided
reasoning, the Model Context Protocol (MCP) provides the actual ``wiring'' for
modern AI agents. Introduced by Anthropic in November 2024~\cite{mcp-spec}, MCP is an
open-source standard designed to solve the ``N-to-M'' integration problem.
Traditionally, if an AI application (host) wanted to connect to ten different
tools (e.g., GitHub, Slack, Google Drive), it had to build ten unique, bespoke
integrations. Conversely, if a new tool wanted to be accessible by multiple LLM
applications, its developers had to write integrations for each one. MCP
standardizes this communication, allowing any compliant host to interact with
any compliant server through a standardized set of primitives. The architecture
of MCP is divided into three distinct roles: the Host, the Client, and the
Server. The Host is the top-level AI environment, such as Claude Desktop or an
AI-powered IDE like Cursor, which orchestrates the overall user experience. The
MCP Client exists within the host and is responsible for maintaining 1:1
connections with one or more MCP servers. The Server is the external service
that provides the actual context, data, or tool-calling capabilities to the
model. Table~\ref{tab:mcp-components} summarizes the responsibilities of each architectural component.

\section{Architectural Participants and Interaction Flows}
\label{sec:mcp-architecture}

The interaction within an MCP ecosystem follows a stateful client-server model,
utilizing JSON-RPC~2.0 as the underlying communication protocol. This choice
ensures that messages are structured, predictable, and easy to parse for both
human-authored code and language models.

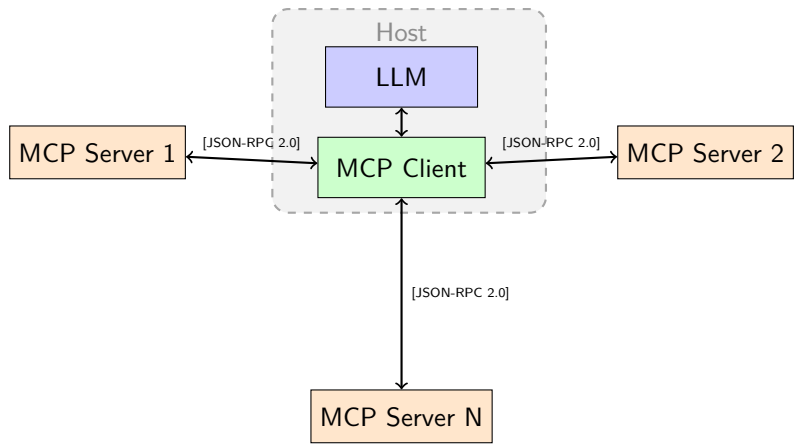
\begin{figure}[ht]
\centering
\begin{tikzpicture}[ node distance=1.5cm]

  \draw[draw=gray!70, fill=gray!10, rounded corners=6pt, thick, dashed]
    (0.8, 1.7) rectangle (4.4, 4.4);
  \node[font=\sffamily, text=gray!90] at (2.5, 4.1) {Host};

  \node[draw, rectangle, fill=blue!20, minimum width=2cm, minimum height=0.8cm] (llm) at (2.5, 3.5) {\textsf{LLM}};

  \node[draw, rectangle, fill=green!20, minimum width=2.2cm, minimum height=0.8cm] (client) at (2.5, 2.3) {\textsf{MCP Client}};

  \node[draw, rectangle, fill=orange!20, minimum width=1.8cm, minimum height=0.7cm] (server1) at (-1.5, 2.5) {\textsf{MCP Server 1}};
  \node[draw, rectangle, fill=orange!20, minimum width=1.8cm, minimum height=0.7cm] (server2) at (6.5, 2.5) {\textsf{MCP Server 2}};
  \node[draw, rectangle, fill=orange!20, minimum width=1.8cm, minimum height=0.7cm] (server3) at (2.5, -1) {\textsf{MCP Server N}};

  \draw[<->,thick] (client) -- (server1) node[midway,above,font=\tiny] {\textsf{[JSON-RPC 2.0]}};
  \draw[<->,thick] (client) -- (server2) node[midway,above,font=\tiny] {\textsf{[JSON-RPC 2.0]}};
  \draw[<->,thick] (client) -- (server3) node[midway,right,font=\tiny] {\textsf{[JSON-RPC 2.0]}};

  \draw[<->,thick] (llm) -- (client);

\end{tikzpicture}
\vspace{0.3cm}
\caption{MCP Architecture: The Host contains an LLM and MCP Client. The Client maintains 1:1 connections to multiple external MCP Servers via JSON-RPC 2.0. This architecture solves the N-to-M integration problem: one host can connect to many servers, and one server can serve many hosts.}
\label{fig:mcp-architecture}
\end{figure}

\begin{table}[ht]
\centering
\small
\caption{MCP architectural components and their responsibilities.}
\label{tab:mcp-components}
\vspace{0.2cm}
\begin{tabularx}{\textwidth}{@{}l X X@{}}
\toprule
\textbf{Component} & \textbf{Responsibility} & \textbf{Relationship} \\
\midrule
MCP Host & Manages the user interface, LLM orchestration, and security policies. & Coordinates multiple clients \\
MCP Client & Handles serialization, connection lifecycle, and session management. & 1:1 with specific server \\
MCP Server & Exposes tools, resources, and prompts from a specific data source. & Provides context to clients \\
\bottomrule
\end{tabularx}
\end{table}

The lifecycle of an MCP connection begins with a capability negotiation phase.
The client initiates an \texttt{initialize} request, presenting its protocol version
and supported features. The server responds with its own capabilities, such as
whether it supports dynamic tool listing or user elicitation. This negotiation
ensures that both parties can communicate effectively despite potential version
mismatches or differences in feature sets. Once initialized, the client sends a
\texttt{notifications/initialized} message, after which the connection is considered
active for data exchange.

\subsection{Standardized Primitives: Tools, Resources, and Prompts}
\label{subsec:mcp-primitives}

At the heart of MCP are three primitives that define the capabilities shared
between servers and clients. These primitives allow developers to expose complex
data and functions in a way that language models can discover and use at runtime
without prior hardcoding.

\begin{description}
\item[Tools:] These are executable functions that the model can invoke to perform
  an action or a side effect, such as creating a GitHub issue or executing a SQL
  query~\cite{gorilla}. Tools accept typed parameters via a JSON Schema and return structured
  results.
\item[Resources:] These are standardized ways for servers to share read-only
  contextual data, such as file contents, database schemas, or logs. Each
  resource is uniquely identified by a URI, and clients can navigate and read
  these resources to ground the LLM's responses in factual data.
\item[Prompts:] These are reusable templates for LLM instructions and workflows.
  They allow servers to provide ``golden'' troubleshooting workflows or standard
  operating procedures directly to the model.
\end{description}

\section{Mapping SGD Principles to the Model Context Protocol}
\label{sec:sgd-mcp-mapping}

The application of SGD concepts to the MCP protocol represents a significant
leap in the practicality of autonomous agents. In the SGD framework, the core
challenge was for a model to interpret an intent and its associated slots based
on a schema. In the MCP world, this translates to the model discovering a ``tool''
and its associated \texttt{inputSchema}.

\begin{figure}[ht]
\centering
\begin{tikzpicture}[
  every node/.style={font=\small\sffamily},
  sgd/.style={draw, rounded corners=4pt, fill=blue!20, minimum width=2.8cm,
              minimum height=0.7cm, align=center, line width=0.6pt},
  mcp/.style={draw, rounded corners=4pt, fill=green!20, minimum width=2.8cm,
              minimum height=0.7cm, align=center, line width=0.6pt},
  arr/.style={->, >=stealth, thick},
  lbl/.style={font=\footnotesize\sffamily, fill=white, inner sep=1.5pt},
]
  \node[font=\small\bfseries\sffamily] at (1, 5.0) {SGD Framework};
  \node[font=\small\bfseries\sffamily] at (7, 5.0) {MCP Protocol};

  \node[sgd] (sgd-intent) at (1, 3.9) {Intent};
  \node[sgd] (sgd-req)    at (1, 2.6) {Required Slots};
  \node[sgd] (sgd-desc)   at (1, 1.3) {Natural Language};
  \node[sgd] (sgd-values) at (1, 0.0) {Slot Values};

  \node[mcp] (mcp-tool)   at (7, 3.9) {Tool};
  \node[mcp] (mcp-req)    at (7, 2.6) {Required Properties};
  \node[mcp] (mcp-desc)   at (7, 1.3) {Description Field};
  \node[mcp] (mcp-enum)   at (7, 0.0) {Enum Constraints};

  \draw[arr] (sgd-intent) -- (mcp-tool)  node[lbl, midway, above] {maps to};
  \draw[arr] (sgd-req)    -- (mcp-req)   node[lbl, midway, above] {maps to};
  \draw[arr] (sgd-desc)   -- (mcp-desc)  node[lbl, midway, above] {maps to};
  \draw[arr] (sgd-values) -- (mcp-enum)  node[lbl, midway, above] {maps to};
\end{tikzpicture}
\vspace{0.3cm}
\caption{Structural mapping between Schema-Guided Dialogue and Model Context Protocol concepts. Each SGD construct has a direct counterpart in MCP, enabling translation of schema-guided reasoning to tool discovery and execution.}
\label{fig:sgd-mcp-mapping}
\end{figure}
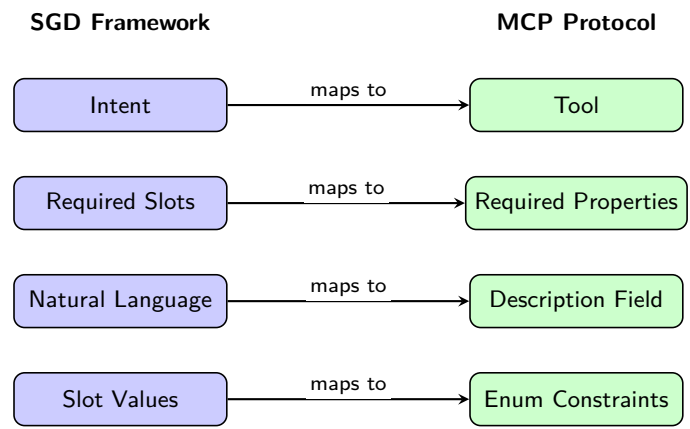

The mapping between these two paradigms is highly deterministic. An SGD ``intent''
is functionally equivalent to an MCP ``tool''. The ``required slots'' in SGD map to
the required properties in an MCP tool's JSON Schema, and the ``natural language
descriptions'' provided in SGD schemas serve exactly the same purpose as the
\texttt{description} field in an MCP tool definition. This metadata is critical because
it allows the model to understand not just what the parameters are, but why and
when they should be used. The structural correspondence is illustrated in Figure~\ref{fig:sgd-mcp-mapping}, with Table~\ref{tab:api-comparison} showing how this paradigm shift reduces integration complexity from $N \times M$ to $N + M$ connections.

\section{The Role of Dynamic Discovery and the USB-C Metaphor}
\label{sec:dynamic-discovery}

MCP is frequently described as the ``USB-C for AI''. Just as a USB-C port allows a
computer to connect to a variety of peripherals---from monitors to storage
drives---without custom drivers for each, MCP allows an AI model to connect to any
data source or tool that implements the protocol. This is achieved through
``dynamic discovery,'' where the model can query the server's \texttt{tools/list} endpoint
at runtime to see available functions. This dynamic nature is a direct
realization of the SGD goal: building a single model that can handle an
ever-increasing number of services without retraining. Instead of relying on
hardcoded API endpoints, the agent uses the schema descriptions to formulate a
JSON-RPC \texttt{tools/call} request with the appropriate arguments. This shift allows
for ``pluggable'' context management, where developers can swap in new tools or
data sources mid-session, and the agent will adapt its behavior based on the new
schema definitions provided.

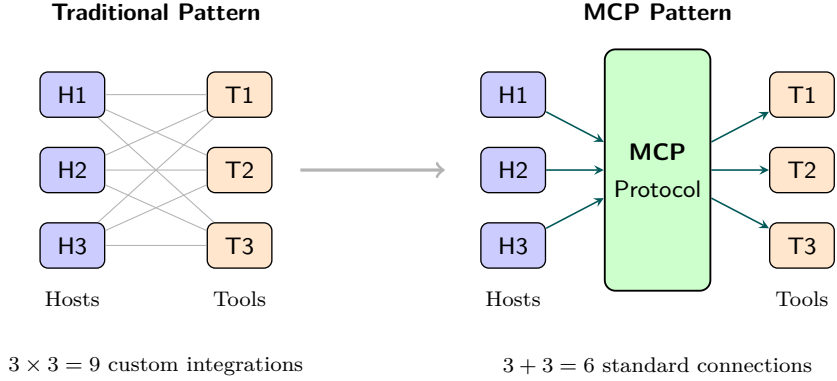
\begin{figure}[ht]
\centering
\begin{tikzpicture}[
  every node/.style={font=\small\sffamily},
  hnode/.style={draw, rounded corners=3pt, fill=blue!20, minimum width=0.85cm,
                minimum height=0.6cm, align=center, line width=0.5pt},
  tnode/.style={draw, rounded corners=3pt, fill=orange!20, minimum width=0.85cm,
                minimum height=0.6cm, align=center, line width=0.5pt},
  pnode/.style={draw, rounded corners=4pt, fill=green!20, minimum width=1.3cm,
                minimum height=3.2cm, align=center, line width=0.7pt},
  conn/.style={thin, gray!60},
  marr/.style={->, >=stealth, semithick, teal!70!black},
]

  \node[font=\small\bfseries\sffamily] at (1.5, 5.6) {Traditional Pattern};

  \node[hnode] (h1) at (0.4, 4.5) {H1};
  \node[hnode] (h2) at (0.4, 3.5) {H2};
  \node[hnode] (h3) at (0.4, 2.5) {H3};
  \node[font=\footnotesize] at (0.4, 1.8) {Hosts};

  \node[tnode] (t1) at (2.6, 4.5) {T1};
  \node[tnode] (t2) at (2.6, 3.5) {T2};
  \node[tnode] (t3) at (2.6, 2.5) {T3};
  \node[font=\footnotesize] at (2.6, 1.8) {Tools};

  \draw[conn] (h1) -- (t1); \draw[conn] (h1) -- (t2); \draw[conn] (h1) -- (t3);
  \draw[conn] (h2) -- (t1); \draw[conn] (h2) -- (t2); \draw[conn] (h2) -- (t3);
  \draw[conn] (h3) -- (t1); \draw[conn] (h3) -- (t2); \draw[conn] (h3) -- (t3);

  \node[font=\footnotesize, align=center] at (1.5, 0.9)
    {$3 \times 3 = 9$ custom integrations};

  \draw[->, very thick, gray!60] (3.4, 3.5) -- (5.3, 3.5);

  \node[font=\small\bfseries\sffamily] at (8.1, 5.6) {MCP Pattern};

  \node[hnode] (h4) at (6.2, 4.5) {H1};
  \node[hnode] (h5) at (6.2, 3.5) {H2};
  \node[hnode] (h6) at (6.2, 2.5) {H3};
  \node[font=\footnotesize] at (6.2, 1.8) {Hosts};

  \node[pnode] (proto) at (8.1, 3.5) {\textbf{MCP}\\[3pt]Protocol};

  \node[tnode] (t4) at (10.0, 4.5) {T1};
  \node[tnode] (t5) at (10.0, 3.5) {T2};
  \node[tnode] (t6) at (10.0, 2.5) {T3};
  \node[font=\footnotesize] at (10.0, 1.8) {Tools};

  \draw[marr] (h4) -- (proto);
  \draw[marr] (h5) -- (proto);
  \draw[marr] (h6) -- (proto);
  \draw[marr] (proto) -- (t4);
  \draw[marr] (proto) -- (t5);
  \draw[marr] (proto) -- (t6);

  \node[font=\footnotesize, align=center] at (8.1, 0.9)
    {$3 + 3 = 6$ standard connections};

\end{tikzpicture}
\vspace{0.3cm}
\caption{Traditional N-to-M Integration Problem vs. MCP/SGD Solution. In traditional patterns, each Host requires custom integrations with each Tool: with 3 hosts and 3 tools, this yields 3×3 = 9 connections, scaling quadratically. MCP solves this by introducing a single standardized protocol: all hosts connect to the same protocol, which connects to all tools, reducing complexity from N×M to N+M connections (3+3 = 6 in this example).}
\label{fig:api-comparison}
\end{figure}

\begin{table}[ht]
\centering
\small
\caption{Comparison of traditional API patterns with MCP/SGD patterns.}
\label{tab:api-comparison}
\vspace{0.2cm}
\begin{tabularx}{\textwidth}{@{}l X X@{}}
\toprule
\textbf{Feature} & \textbf{Traditional API Pattern} & \textbf{MCP / SGD Pattern} \\
\midrule
Integration & Bespoke, hardcoded connectors & Standardized, schema-guided \\
Discovery & Developer-defined at build time & Machine-discoverable at runtime \\
Context & Static, training-time knowledge & Dynamic, retrieved from live sources \\
Interaction & One-off request-response & Stateful, multi-turn orchestration \\
Scalability & Linear (N integrations for N tools) & Constant (One protocol for all tools) \\
\bottomrule
\end{tabularx}
\end{table}

\section{State Tracking and Multi-Turn Reasoning in Agentic Systems}
\label{sec:state-tracking}

In the research on schema-guided dialogue, Dialogue State Tracking (DST) is the
process of building a representation of the user's goal by aggregating
information across turns~\cite{dstc}. In the context of MCP and industrial AI agents, this
has evolved into more sophisticated ``Context Management'' and ``Session
Isolation''. Because real-world tasks often involve long horizons---defined as
tasks requiring more than ten interdependent reasoning and action
steps---maintaining a coherent state is critical~\cite{compass}.

Persistent storage solutions, such as distributed databases or cloud-based storage systems, are often utilized in
enterprise MCP implementations to store serialized conversation threads and
workflow checkpoints~\cite{mcp-microsoft}. This allows the agent to maintain continuity even across
restarts or multi-day interactions. The ``state'' of the agent is no longer just a
collection of slot values but a comprehensive trace of execution, including
intermediate tool outputs, reflections on past errors, and revised plans.

\section{The COMPASS Architecture for Long-Horizon Tasks}
\label{sec:compass}

As agents tackle increasingly complex tasks, a single-agent architecture often
fails due to ``context exhaustion,'' where the growing execution trace obscures
critical information. To mitigate this, the Context-Organized Multi-Agent
Planning and Strategy System (COMPASS) was proposed~\cite{compass}. COMPASS is a hierarchical
framework that decouples different aspects of reasoning to ensure strategic
coherence over long horizons.

The COMPASS architecture consists of three specialized components:

\begin{enumerate}
\item \emph{Context Manager}: Organizes and synthesizes the execution history. Instead
  of passing the entire, noisy history to the Main Agent, it produces optimized
  ``Context Briefs'' that highlight the task goal, verified facts, and
  prioritized next steps.
\item \emph{Main Agent}: Responsible for tactical execution. It operates in a ReAct loop
  (thought-action-observation)~\cite{react, chain-of-thought}, performing specific tool interactions based on
  a ``refreshed'' context brief.
\item \emph{Meta-Thinker}: Acts as a strategic overseer. It monitors the agent's
  trajectory to detect anomalies like strategy drift or repeated failures,
  issuing interventions such as ``Pivot'' or ``Verify''.
\end{enumerate}

\begin{figure}[ht]
\centering
\begin{tikzpicture}[
  every node/.style={font=\small\sffamily},
  box/.style={draw, rounded corners=4pt, minimum width=2.8cm, minimum height=0.9cm,
              align=center, line width=0.6pt},
  subbox/.style={draw, rounded corners=4pt, minimum width=2.8cm, minimum height=0.75cm,
                 align=center, line width=0.6pt},
  arr/.style={->, >=stealth, semithick},
  darr/.style={->, >=stealth, semithick, dashed},
  lbl/.style={font=\footnotesize\sffamily, fill=white, inner sep=1.5pt},
]
  \node[box, fill=gray!20, minimum width=4.2cm] (hist) at (4.5, 5.8)
    {Execution History};

  \node[box, fill=blue!20]   (cm) at (0.8, 3.6) {Context\\Manager};
  \node[box, fill=green!20]  (ma) at (4.5, 3.6) {Main Agent};
  \node[box, fill=orange!20] (mt) at (8.2, 3.6) {Meta-Thinker};

  \node[subbox, fill=blue!8]   (brief)   at (0.8, 1.8) {Context Brief};
  \node[subbox, fill=green!8]  (tools)   at (4.5, 1.8) {Tool Calls};
  \node[subbox, fill=orange!8] (monitor) at (8.2, 1.8) {Monitor \&\\Intervene};

  \draw[arr] (hist.west)  -- (cm.north);
  \draw[arr] (hist.south) -- (ma.north);
  \draw[arr] (hist.east)  -- (mt.north);

  \draw[arr] (cm) -- (brief);
  \draw[arr] (ma) -- (tools);
  \draw[arr] (mt) -- (monitor);

  \draw[darr] (brief.north east) to[out=55, in=225]
    node[lbl, pos=0.45] {refreshes} (ma.south west);

  \draw[darr] (monitor.north west) to[out=125, in=-45]
    node[lbl, pos=0.45] {pivot/verify} (ma.south east);

  \draw[arr] (tools.east) -- ++(0.5, 0) -- ++(0, 4.0) -- (hist.east);
  \node[lbl] at (6.55, 3.8) {feedback};

\end{tikzpicture}
\vspace{0.3cm}
\caption{COMPASS Architecture: A hierarchical multi-agent system for long-horizon task execution. The Context Manager synthesizes execution history into optimized briefs. The Main Agent executes tools in a ReAct loop using context briefs. The Meta-Thinker monitors trajectory and issues strategic interventions. Information flows in both directions: briefs refresh the agent, tool results feed back into execution history.}
\label{fig:compass-architecture}
\end{figure}
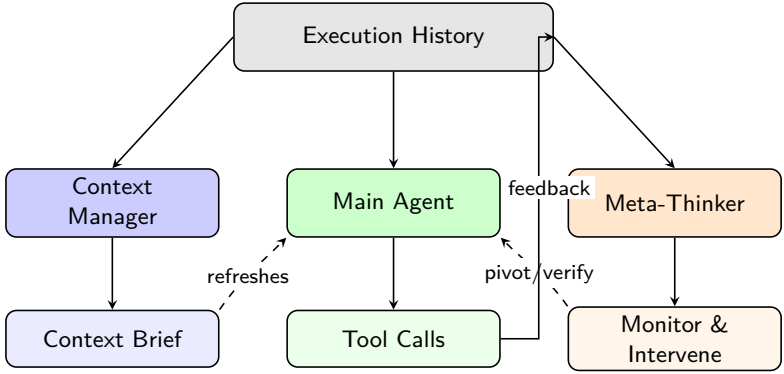

This multi-agent approach mirrors the multi-pass encoders used in FastSGT~\cite{fastsgd} and
other SGD models, where information is distilled and processed in layers to
improve understanding and efficiency. By separating strategic memory from local
tool execution, preliminary results suggest that COMPASS can improve accuracy on specific complex task benchmarks, with reported gains up to 20\% on controlled settings~\cite{compass}---though generalization across diverse domains remains an open question.

\section{Benchmarking Agentic Performance in MCP Ecosystems}
\label{sec:benchmarking}

The rise of the MCP protocol has necessitated new methods for evaluating the
``tool-use'' capabilities of language models. Traditional benchmarks often rely on
static text reasoning, but the ``Agent Web'' requires testing the model's ability
to interact with real-world servers, handle unfamiliar schemas, and recover from
execution errors.

\subsection{MCP-Universe and Real-World Evaluation}
\label{subsec:mcp-universe}

MCP-Universe is the first comprehensive benchmark designed to evaluate LLMs on
realistic tasks through interaction with actual MCP servers. It covers domains
like location navigation, financial analysis, repository management, and 3D
design using 11 real-world MCP servers~\cite{mcp-universe}. The evaluation is execution-driven,
meaning it assesses whether the model successfully completed a task (e.g.,
``count the rows in a MySQL table'') rather than just checking its text output
against a reference.

Findings from the MCP-Universe benchmark reveal significant performance gaps,
even in frontier models. For instance, top-tier models like GPT-5-High achieved
success rates of only 44.16\%, while Grok-4 managed 33.33\%~\cite{mcp-universe}. These results
highlight the ``unknown-tools'' challenge: even if a model is highly intelligent,
it may struggle to use a specific MCP server correctly if the schema is
ambiguous or if the task requires navigating a large space of available tools. Table~\ref{tab:benchmarks} compares the scope and evaluation methods of multiple MCP benchmarking frameworks.

\subsection{MCPAgentBench and Tool Discrimination}
\label{subsec:mcpagentbench}

MCPAgentBench provides a complementary evaluation focus, specifically targeting
the efficiency with which agents select and invoke tools~\cite{mcpagentbench}. It employs a sandbox
environment with distractors---irrelevant tool definitions included in the
prompt---to test the agent's ``discrimination'' and ``anti-interference'' abilities.
This ``needle in a haystack'' scenario is representative of real-world enterprise
environments where an agent might have access to thousands of tools across
multiple servers.

\begin{table}[ht]
\centering
\small
\caption{Comparison of MCP benchmarking frameworks.}
\label{tab:benchmarks}
\vspace{0.2cm}
\begin{tabularx}{\textwidth}{l X X X}
\toprule
\textbf{Benchmark} & \textbf{Scope} & \textbf{Evaluation Method} & \textbf{Key Insight} \\
\midrule
MCP-Universe & 6 domains, 231 tasks & Execution-driven (Dynamic) & Frontier models still struggle with long-horizon reasoning. \\
MCPAgentBench & 9,714 servers, 20,000 tools & Sandbox with distractors & Tool selection efficiency is a major bottleneck. \\
ToolACE-MCP & History-aware routing & Routing accuracy & History is crucial for selecting the right tool in sequences. \\
MCPMark & Real-world workflows & Execution turns / Task & Agents require an average of 16.2 turns per task. \\
\bottomrule
\end{tabularx}
\end{table}

\section{Optimizing for Efficiency: Token Bloat \& Discovery Strategies}
\label{sec:optimization}

A significant technical hurdle in scaling schema-guided MCP systems is ``context
bloat'' or ``token bloat.'' When an agent has access to dozens of tools, the JSON
schemas defining those tools can take up 40--50\% of the model's available context
window~\cite{token-bloat}. This not only increases cost and latency but also leads to ``LLM
confusion,'' where the model becomes less accurate because it is overwhelmed by
irrelevant documentation.

To address this, researchers have developed ``active'' agent frameworks like
MCP-Zero~\cite{mcp-zero}. Instead of injecting all tool schemas upfront, MCP-Zero restores
autonomy to the model, allowing it to actively identify capability gaps and
request only the specific tools it needs on-demand. This is achieved through a
hierarchical semantic routing algorithm that first filters candidate servers and
then ranks tools based on semantic similarity to the agent's current goal.

\subsection{Token Reduction and Progressive Disclosure}
\label{subsec:token-reduction}

Other strategies for reducing bloat include the ``describe\_tools'' pattern~\cite{token-reduction}. In
this approach, the model is first provided with a high-level overview of tool
categories. Only when the model decides it needs to use a specific tool (e.g., a
GitHub tool) does the system load the full, detailed JSON schema for that
specific function. As shown in Table~\ref{tab:token-strategies}, this progressive disclosure approach offers dramatic token reductions at the cost of additional tool calls.

\begin{table}[ht]
\centering
\small
\caption{Token optimization strategies for MCP systems.}
\label{tab:token-strategies}
\vspace{0.2cm}
\begin{tabularx}{\textwidth}{l X X X}
\toprule
\textbf{Strategy} & \textbf{Mechanism} & \textbf{Impact on Tokens} & \textbf{Impact on Latency} \\
\midrule
Static Injection & Inject all schemas upfront & High bloat, linear growth & Low (no extra discovery steps) \\
Semantic Search & Find tools via embeddings & Reduced bloat & Moderate (retrieval overhead) \\
Progressive Disclosure & Summaries first, details later & Massive reduction (90\%+) & Higher (2--3x more tool calls) \\
MCP-Zero & Active discovery on-demand & 98\% reduction & Variable (depends on search path) \\
\bottomrule
\end{tabularx}
\end{table}

Experimental results show that while progressive disclosure and active discovery
patterns increase the number of tool calls (as the agent must first ``search'' for
a tool), the total token consumption drops by an average of 96\% for simple
tasks~\cite{mcp-zero, token-reduction}. This trade-off is often acceptable in large-scale systems where the
alternative is exceeding the context window entirely.

\section{Security, Trust, \& Safety in the MCP Ecosystem}
\label{sec:security}

The Model Context Protocol enables powerful capabilities through arbitrary data
access and code execution, which necessitates a rigorous security framework.
Unlike traditional APIs where security is often handled at the network
perimeter, MCP security is centered on the principle of ``human-in-the-loop'' and
explicit user consent~\cite{mcp-spec, mcp-security}.

\subsection{Security Principles \& Attack Vectors}
\label{subsec:security-vectors}

The protocol defines several key security principles that all implementors must
address. First, user consent is mandatory; agents must not perform
actions---especially transactional ones---without explicit approval through an
authorization UI. Second, tool safety is prioritized by treating tool
descriptions as ``untrusted'' input. If a malicious server provides a tool
description that tricks the model into executing an exploit, the host
application is responsible for enforcing guardrails.

A significant risk in this ecosystem is the ``Tool Poisoning Attack'' (TPA)~\cite{mcp-csa, mcp-security}, where
malicious instructions are embedded within tool descriptions. Because LLMs are
designed to follow instructions, a cleverly worded tool description could
persuade the model to ignore user preferences or exfiltrate sensitive data. To
counter this, organizations are encouraged to apply supply-chain controls,
review tool definitions like source code, and use identity-aware access layers
to govern model interactions. Table~\ref{tab:security} summarizes the attack vectors and mitigation strategies in MCP ecosystems.

\begin{table}[h!]
\centering
\small
\caption{Attack vectors and mitigation strategies in MCP ecosystems.}
\label{tab:security}
\vspace{0.2cm}
\begin{tabularx}{\textwidth}{l X X}
\toprule
\textbf{Attack Type} & \textbf{Description} & \textbf{Mitigation Strategy} \\
\midrule
Prompt Injection & Malicious input tricks LLM into unsafe tool use. & Input sanitization, policy enforcement. \\
Tool Poisoning & Hidden instructions in tool metadata manipulate agent. & Treat metadata as untrusted, integrity checks. \\
Orchestration Injection & Hostile instructions in retrieved data steer model. & Segregate untrusted content, sandbox execution. \\
Lookalike Tools & Fake tools impersonate legitimate ones to capture data. & Maintain an allowlist of approved servers. \\
Privilege Escalation & Server uses broad privileges instead of user-bound ones. & Enforce user-bound scopes and token validation. \\
\bottomrule
\end{tabularx}
\end{table}

\section{Network Management and Intent-Based Troubleshooting}
\label{sec:network-management}

In the domain of network management, researchers have proposed mapping MCP
primitives to the network management plane to allow operators to troubleshoot
infrastructure using natural language~\cite{mcp-ietf}. In this setup, a router or switch acts as
an MCP server, while a network controller acts as the MCP client. Table~\ref{tab:network} illustrates how MCP primitives map to network management concepts.

\begin{table}[ht]
\centering
\small
\caption{Mapping MCP primitives to network management equivalents.}
\label{tab:network}
\vspace{0.2cm}
\begin{tabularx}{\textwidth}{l X X}
\toprule
\textbf{MCP Primitive} & \textbf{Network Management Equivalent} & \textbf{Example} \\
\midrule
Resource & Read-only YANG datastores, syslogs & \texttt{mcp://router1/ietf-interfaces} \\
Tool & Idempotent actions, RPC operations & ping, traceroute, clear counters \\
Prompt & Golden troubleshooting workflows & ``Diagnose why BGP neighbor is down'' \\
\bottomrule
\end{tabularx}
\end{table}

This mapping allows a controller's LLM to deduce the required sequence of tool
calls---such as running a ping and then reading an interface state---to diagnose an
issue and recommend a configuration change, all while ensuring that high-impact
actions are gated by human approval.

\section{Mathematical Foundations for Ensembled Agent Decisions}
\label{sec:mathematical-foundations}

To improve the accuracy of tool selection and parameter generation in complex
tasks, especially when ground-truth data is unavailable, agentic systems often
utilize ensemble voting mechanisms. This is particularly relevant in
``Text-to-SQL'' or ``Schema Linking'' tasks where multiple LLM ``experts'' might
propose different solutions.

The Weighted Majority Algorithm (WMA) is a common technique used to coordinate
these expert predictions~\cite{wma-littlestone, wma-sql}. Each expert agent $i$ is assigned a weight $w_i$,
initially uniform (e.g., $w_i = 1.0$). Given a set of candidate solutions $\{s\}$
proposed by different experts, the system computes the total weight for each
candidate:

\begin{equation}
W(s) = \sum_{i : f_i = s} w_i
\end{equation}

The solution with the highest total weight is selected. If supervision---such as
an execution result or gold label---is available, the weights of experts who
provided incorrect answers are updated with a multiplicative penalty $(1 - \epsilon)$:

\begin{equation}
w_i \leftarrow w_i \cdot (1 - \epsilon), \quad \text{if } f_i \neq s^*
\end{equation}

This approach minimizes the ``mistake bound'' of the system. To further introduce
diversity and prevent the system from getting stuck in local optima, the
Randomized Weighted Majority Algorithm (RWMA)~\cite{wma-littlestone} can be used, where the final
solution is sampled probabilistically based on the normalized weights:

\begin{equation}
P(s) = \frac{W(s)}{\sum_{s'} W(s')}
\end{equation}

These mathematical frameworks ensure that schema-guided systems can become
self-improving, gradually learning which agents are more reliable for specific
domains or toolsets.

\section{Schema Design Principles for LLM-Native Interoperability}
\label{sec:schema-design}

The convergence of SGD and MCP reveals a fundamental paradox: while the protocol standardizes how schemas are exchanged, it says little about what makes a schema effective for LLM consumption. Traditional JSON Schema was designed for human developers and static validators---not for language models that must infer intent from natural language descriptions. As Software~3.0 matures, schema quality becomes the primary determinant of agent reliability~\cite{parse}. Research on instruction tuning~\cite{instruction-tuning-gpt4} and tool learning with foundation models~\cite{tool-learning-fm} demonstrates that semantic richness in descriptions significantly impacts model performance.

By analyzing the parallel evolution of SGD and MCP, we identify five foundational principles for LLM-native schema design. These principles are not derived from first principles but rather excavated from the specific design choices each framework made to address the same underlying problem: how do we enable a language model to dynamically discover and invoke external services?

\subsection{Semantic Completeness Over Syntactic Precision}
\label{subsec:semantic-completeness}

\begin{description}
\item[SGD Perspective:] The breakthrough of the Schema-Guided Dialogue framework was recognizing that natural language descriptions of intents and slots enabled zero-shot generalization across domains~\cite{maml, icl-foundations}. The challenge of understanding user intent and extracting relevant parameters has deep roots in dialogue systems research~\cite{intent-slot-joint, spoken-language-understanding, intent-detection-survey}. A \texttt{GetWeather} intent required not just the function signature but a description explaining \emph{when} to call it and \emph{why} it matters: ``Retrieves current temperature and conditions for a specified location.'' Similarly, a slot like \texttt{departure\_city} needed semantic grounding: ``The IATA code for the departure airport (e.g., ZRH, JFK)'' rather than just \texttt{string}. This semantic richness is what enabled a single model to generalize to entirely new APIs during inference, leveraging in-context learning mechanisms~\cite{incontext-learning-review} to adapt to novel task structures.

\item[MCP Perspective:] In the Model Context Protocol, tool descriptions are the \emph{primary discovery mechanism} at runtime. An \texttt{inputSchema} alone---specifying that a parameter is a string or integer---is insufficient. The \texttt{description} field drives tool selection and agent reasoning about intent. A GitHub tool must be described as ``Creates an issue in the specified repository with a title and body'' rather than just mapping to a REST endpoint. MCP servers that provide only minimal descriptions see poor tool utilization because agents cannot reason about appropriate invocation contexts. This parallels the challenge of intent detection at scale: when descriptions are vague or syntactically focused, agents struggle to match user needs with available tools.

\item[Convergence Insight:] Both paradigms independently learned that LLMs need \emph{why} and \emph{when}, not just \emph{what} and \emph{how}. This represents a fundamental departure from REST API documentation, which is authored for human developers who can consult specification pages and infer context. Schema design for LLM agents must treat description fields as first-class, semantically rich artifacts---grounded in the research on few-shot learning and intent detection that shows natural language is the primary signal for agent reasoning.
\end{description}

\subsection{Explicit Action Boundaries}
\label{subsec:explicit-action-boundaries}

\begin{description}
\item[SGD Perspective:] The SGD framework included an explicit \texttt{is\_transactional} field that flagged state-changing operations. Services marked as transactional required user confirmation before execution (e.g., booking a flight or processing a payment). The distinction between read operations (search queries) and write operations (transactions) was encoded in the schema, allowing dialogue systems to enforce guardrails without requiring post-hoc reasoning.

\item[MCP Perspective:] The Model Context Protocol notably lacks explicit transactional flags---this is a gap in the current specification. Instead, MCP relies on naming conventions (\texttt{get\_}, \texttt{create\_}, \texttt{delete\_}) and host-level guardrails for human-in-the-loop approval on destructive operations. Your banking case study illustrated the importance of ``deterministic workflow control that gates high-risk actions'': without explicit action boundary declarations, orchestrating multi-step tool chains becomes unreliable. This need is empirically grounded: analysis off our internal MCP orchestrator revealed over >1000 tool-to-tool dependencies across the tool ecosystem. Our eco-system consists of >10 agents, each an expert on its domain, jointly working together. It validates the claim, that at scale, implicit action boundaries become untenable.

The critical gap: MCP should standardize a formal mechanism for declaring tool dependencies and action boundaries in schema descriptions---making dependency metadata machine-readable and formally accessible rather than implicit in tool names or agent reasoning. Without such standardization, agents face exponential reasoning burden and increased failure rates in multi-step orchestration.

\item[Convergence Insight:] SGD had this principle explicitly; MCP currently relies on conventions and host-level safeguards. This convergence analysis reveals a critical standardization gap: MCP must formalize explicit action boundary declarations in its schema specification. This includes an \texttt{actionType} field that distinguishes read, write, and destructive operations, and a structured dependency declaration mechanism that makes tool-to-tool relationships formally accessible. Such standardization is essential for reliable multi-step orchestration at scale---transforming implicit reasoning into explicit, verifiable constraints. In a productive federated gateway implementation, explicit \texttt{actionType} declarations in tool schemas enforce action boundaries independently of agent interpretation, preventing guardrail bypass even under schema misreading. This validation approach is detailed in forthcoming technical work.
\end{description}

\subsection{Failure Mode Documentation}
\label{subsec:failure-modes}

\begin{description}
\item[SGD Perspective:] The SGD simulator, trained on diverse dialogue trajectories, implicitly learned how to handle failures: slots that could not be filled, services temporarily unavailable, or user requests that exceeded API constraints. Dialogue state tracking enabled recovery strategies (retry, clarify with user, suggest alternative). However, these recovery strategies were learned implicitly from training data rather than explicitly documented in the schema.

\item[MCP Perspective:] Modern MCP tools return structured results with success/error fields, but error semantics are ad-hoc. There is no standardized way for a tool to communicate ``this failed because the resource doesn't exist'' versus ``this failed because rate limits were exceeded'' versus ``this failed because authentication credentials were invalid.'' Agents lack explicit guidance on recovery strategies: should they retry? switch to an alternative tool? ask the user for clarification? The analogy to OpenAPI's response codes (200, 404, 500) is instructive---those codes provide semantic information to \emph{human} developers; LLM agents need the equivalent.

\item[Convergence Insight:] Neither framework explicitly documented failure modes as first-class schema elements. This is a novel gap your work identifies: failure mode documentation should enumerate expected error conditions and recovery strategies, analogous to OpenAPI's \texttt{responses} structure but semantically grounded for LLM reasoning. This principle says: treat error pathways as important as success pathways.
\end{description}

\subsection{Progressive Disclosure Compatibility}
\label{subsec:progressive-disclosure}

\begin{description}
\item[SGD Perspective:] The SGD framework provided all schema details upfront to the model. In 2019, when SGD was designed, token constraints were not a limiting factor, and schema size was not a bottleneck: even across 20 domains with hundreds of slots, the schema fit comfortably in context. Progressive disclosure was not necessary. Research on few-shot learning~\cite{maml, icl-foundations} and in-context learning~\cite{incontext-learning-review} had not yet revealed the cost of verbose context windows.

\item[MCP Perspective:] As detailed empirically in Section~\ref{subsec:token-reduction}, progressive disclosure patterns dramatically reduce token consumption while maintaining reasoning quality. The principle emerges as a fundamental architectural constraint: schemas must support explicit two-level design, where concise summaries enable tool discovery while detailed specifications are deferred until needed. This is not optional optimization---it is a requirement for scaling beyond single-domain systems.

\item[Convergence Insight:] This principle is pure MCP innovation, backward-applied as a foundational principle. SGD didn't face scaling pressures; MCP does. The design requirement is clear: schemas must be structurally capable of progressive disclosure. This means tool descriptions must separate discovery-time metadata (name, category, one-line summary) from invocation-time details (full parameter documentation). Without this separation at the schema level, systems cannot implement the empirically-validated patterns documented in Section~\ref{subsec:token-reduction}.
\end{description}

\subsection{Inter-Tool Relationship Declaration}
\label{subsec:inter-tool-relationships}

\begin{description}
\item[SGD Perspective:] Multi-domain conversations required tracking dependencies. A frame-based representation captured that hotel booking depends on the destination city extracted from a preceding flight search. These relationships were implicit in the dialogue flow and dialogue state representation, not explicitly encoded in the schema.

\item[MCP Perspective:] Tools are discovered and invoked independently; there is no explicit dependency graph. Agents must infer ``call authenticate before list\_repos'' or ``the ID returned by create\_order is required by confirm\_order.'' Sequential tool calls are common (MCPAgentBench reports 16.2 average turns per task), and relationship inference adds cognitive burden to agent reasoning. At scale, relationship metadata matters significantly for orchestration. This challenge mirrors knowledge graph research~\cite{knowledge-graphs-review, knowledge-graph-refinement}, where explicit relationship representation is critical for reasoning systems.

\item[Convergence Insight:] SGD tracked state across conversation turns; MCP needs explicit dependency metadata in the schema. This principle says: make relationships first-class schema elements (e.g., ``requires: [authenticate]'' or ``output.id → tool\_Y.input.order\_id''), reducing reasoning burden and improving multi-step completion rates. Drawing inspiration from knowledge graph design patterns~\cite{relation-extraction-knowledge-graphs}, schemas should encode not just tool definitions but the relational structure that connects them.
\end{description}

\subsection{Meta-Analysis: What These Principles Reveal}
\label{subsec:meta-analysis}

The five principles split into a revealing pattern:

\begin{itemize}
\item \textbf{Vindication of SGD's design:} Principles 1 and 2 (semantic completeness and explicit action boundaries) validate SGD's original architectural choices. These principles worked in SGD and should be inherited by MCP.

\item \textbf{Unaddressed gaps:} Principles 3 and 5 (failure modes and inter-tool relationships) identify gaps neither framework explicitly addressed. Neither SGD nor MCP made these first-class schema concerns, yet both encounter them in practice.

\item \textbf{Production scaling insight:} Principle 4 (progressive disclosure) emerges purely from MCP's scaling experience. This is a lesson that SGD's bounded research setting never faced.
\end{itemize}

The deeper insight: \textbf{schema design is not a solved problem}. It is an emerging discipline that requires explicit, principled attention. These five principles draw on decades of research---from dialogue systems and intent detection~\cite{intent-slot-joint, spoken-language-understanding} through few-shot learning and in-context adaptation~\cite{maml, icl-foundations, incontext-learning-review} to knowledge graph design and relational reasoning~\cite{knowledge-graphs-review, relation-extraction-knowledge-graphs}. Yet they address a fundamentally new challenge: how to package semantic knowledge for machine reasoning agents that operate under strict token constraints while navigating complex tool ecosystems at scale.

These five principles suggest that future MCP implementations should treat schema authoring as consequential as API design was in the REST era---but optimized for a machine reader that reasons rather than parses. Organizations adopting MCP should invest in schema design as a first-class engineering concern, recognizing that agent reliability is fundamentally bounded by schema quality. The convergence of SGD and MCP reveals that this investment is not incremental refinement but foundational innovation: we are learning to speak to machines in their native language---semantic, relational, context-aware---rather than expecting them to translate human documentation.

\section{The Future of Agentic Systems and Software 3.0}
\label{sec:future-software3}

The trajectory of schema-guided dialogue and the Model Context Protocol points
toward a fundamental shift in how software is built and consumed. This is the
era of ``Software~3.0''~\cite{software3-karpathy}\footnote{Software~3.0 extends Karpathy's Software~2.0 paradigm by elevating schema-driven agent orchestration as the primary runtime abstraction.}, where the primary consumer of an API is no longer a human developer writing deterministic code, but an autonomous agent interpreting
dynamic schemas. Where Software~2.0 replaced hand-coded logic with learned weights,
Software~3.0 replaces static weights with dynamic, schema-driven agent orchestration.

While Section~\ref{sec:schema-design} establishes the principles that human schema designers should follow, the practical challenge of scaling schema quality points toward automated solutions. The gap between idealized design principles and real-world schema complexity suggests that future MCP ecosystems will require tooling to bridge this gap. Systems like PARSE (Parameter Automated Refinement and Schema Extraction)~\cite{parse} represent a new class of ``schema optimization'' tools: given an existing (possibly suboptimal) schema, these systems automatically refine descriptions, detect field relationships, and restructure constraints to optimize for LLM consumption. Rather than requiring every schema author to master the five principles manually, organizations can leverage automated feedback loops where PARSE analyzes agent failures, identifies ambiguous descriptions, and recommends schema refinements. This shift from manual schema curation to automated schema optimization is analogous to the evolution from hand-written code to compiler-driven code generation---the principles remain important, but tooling makes them scalable and accessible to developers who may not be schema experts.

\section{Conclusion}
\label{sec:conclusion}

This paper has traced the convergence of two powerful paradigms---Schema-Guided Dialogue and the Model Context Protocol---that together form the foundation of Software~3.0~\cite{software3-karpathy}. We have shown that this convergence is not incidental but reflects a fundamental shift in how autonomous agents discover, interpret, and utilize external services.

\subsection{Summary of Key Contributions}
\label{subsec:key-contributions}

The SGD framework solved the ``ontology bottleneck'' by demonstrating that a single model could generalize zero-shot across disparate APIs when provided with natural language schema descriptions. This research breakthrough established the theoretical foundations for dynamic service integration. The Model Context Protocol operationalized these principles into a practical, standardized protocol that any LLM host can use to connect with any compliant server. The structural mapping between SGD intents and MCP tools, between SGD slots and MCP input schemas, reveals that the two paradigms are fundamentally aligned---representing different points on the same spectrum of agent-service interaction.

Building on this foundation, we examined how modern systems like COMPASS apply hierarchical multi-agent architectures to overcome context exhaustion in long-horizon tasks. We reviewed benchmarks (MCP-Universe, MCPAgentBench, ToolACE-MCP) that reveal performance gaps even in frontier models, highlighting that tool selection and schema interpretation remain open challenges. We discussed optimization strategies---from token reduction via progressive disclosure to active discovery patterns---that make agentic systems practical at scale. We analyzed the security landscape, identifying tool poisoning attacks and other vectors that require human-in-the-loop oversight and supply-chain controls.

Critically, this paper proposes five foundational principles for designing schemas that serve LLM agents effectively. These principles---\textit{semantic completeness over syntactic precision}, \textit{explicit action boundaries}, \textit{failure mode documentation}, \textit{progressive disclosure compatibility}, and \textit{inter-tool relationship declaration}---address the gap between what MCP standardizes (how schemas are exchanged) and what remains unspecified (what makes a schema effective for LLM consumption). These principles shift schema authoring from an afterthought to a first-class engineering discipline, recognizing that agent reliability is fundamentally bounded by schema quality.

\subsection{Open Challenges and Future Directions}
\label{subsec:open-challenges}

Several challenges remain unresolved. First, empirical validation of the schema design principles proposed in Section~\ref{sec:schema-design} across diverse domains and use cases remains incomplete. While the five principles are grounded in research, systematic ablation studies are needed to understand which principles matter most for specific agent architectures and task types. Second, schema versioning and backward compatibility in live MCP deployments is unexplored territory. As APIs evolve, how should schema changes be communicated to agents without breaking existing integrations? Third, the interplay between model capability, schema quality, and prompt engineering is poorly understood. Ablation studies that isolate these factors would clarify what aspects of the agent architecture matter most.

A critical frontier is scaling beyond token limits. Current strategies trade computation for context compression, but fundamental questions remain about how to maintain coherent agent reasoning across very long horizon tasks spanning days or weeks. Persistent state management and hierarchical reasoning architectures like COMPASS offer promise, but more research is needed on their scalability and reliability.

\subsection{Toward a Universal AI Substrate}
\label{subsec:universal-substrate}

The convergence of SGD and MCP represents the ``connective tissue'' for the emerging AI-powered enterprise. By standardizing how context is packaged, persisted, and passed to models, these protocols enable memory, grounding, and purpose within every session. They move beyond the era of brittle fine-tuning and ad-hoc prompt stuffing into a world of structured, auditable, and scalable AI connectivity.

As the ecosystem matures, the Model Context Protocol is positioned to become as fundamental as HTTP or USB~\cite{mcp-spec}. It is not merely a way to call a tool; it is a universal standard for interoperability that ensures AI agents remain accurate, secure, and easy to integrate across the vast landscape of human data and services. The synthesis of SGD's theoretical insights and MCP's practical implementation ensures that the next generation of autonomous agents will not just reason, but act---transforming from simple chat interfaces into intelligent agents that can navigate the digital world with precision, transparency, and safety.

In a world where APIs proliferate and integrate at ever-increasing speed, agents that can dynamically discover and reason about new services represent a fundamental shift in software architecture. Software~3.0 is not a distant vision; it is emerging today through the convergence documented in this paper.

\section*{Acknowledgments}

The author gratefully thanks Monika Kaiser Zengaffinen and SBB-IT for the support and institutional backing that made this research possible. This work reflects the conviction that rigorous, principled approaches to AI safety and auditability are essential---values that SBB exemplifies in its mission-critical operations.

\bibliographystyle{plain}
\bibliography{schlapbach-master}

\end{document}